\documentclass[10pt,twocolumn,letterpaper]{article}

\usepackage{cvpr}              
\usepackage{bbm}
\usepackage{verbatim}

\usepackage[misc]{ifsym}

\usepackage{float}
\usepackage{graphicx}
\usepackage{epstopdf}
\usepackage{amsmath}
\usepackage{amssymb}
\usepackage{pifont}
\newcommand{\cmark}{\ding{51}}%
\newcommand{\xmark}{\ding{55}}%
\usepackage{algorithm}
\usepackage{algorithmic}
\usepackage{booktabs}
\usepackage{bbold}
\usepackage{multirow}
\usepackage{colortbl}
\usepackage{xcolor}
\usepackage{bm}
\usepackage{soul}
	\definecolor{airforceblue}{rgb}{0.36, 0.54, 0.66}

\usepackage[pagebackref,breaklinks,colorlinks]{hyperref}

\usepackage[capitalize]{cleveref}
\crefname{section}{Sec.}{Secs.}
\Crefname{section}{Section}{Sections}
\Crefname{table}{Table}{Tables}
\crefname{table}{Tab.}{Tabs.}

\def\cvprPaperID{2809} 
\def\confName{CVPR}
\def\confYear{2022}

\begin{document}

\title{Consensus Synergizes with Memory: \\ A Simple Approach for Anomaly Segmentation in Urban Scenes}

\author{Jiazhong Cen$^1$, Zenkun Jiang$^1$, Lingxi Xie$^2$, Qi Tian$^2$, Xiaokang Yang$^1$, Wei Shen$^{1{(\textrm{\Letter})}}$\\
$^1$MoE Key Lab of Artificial Intelligence, AI Institute, Shanghai Jiao Tong University\\
$^2$Huawei Inc.\\
}


\twocolumn[{%
\renewcommand\twocolumn[1][]{#1}%
\maketitle
\begin{center}
    \vspace{-6mm}
  \centering
    \includegraphics[width=1.0\textwidth]{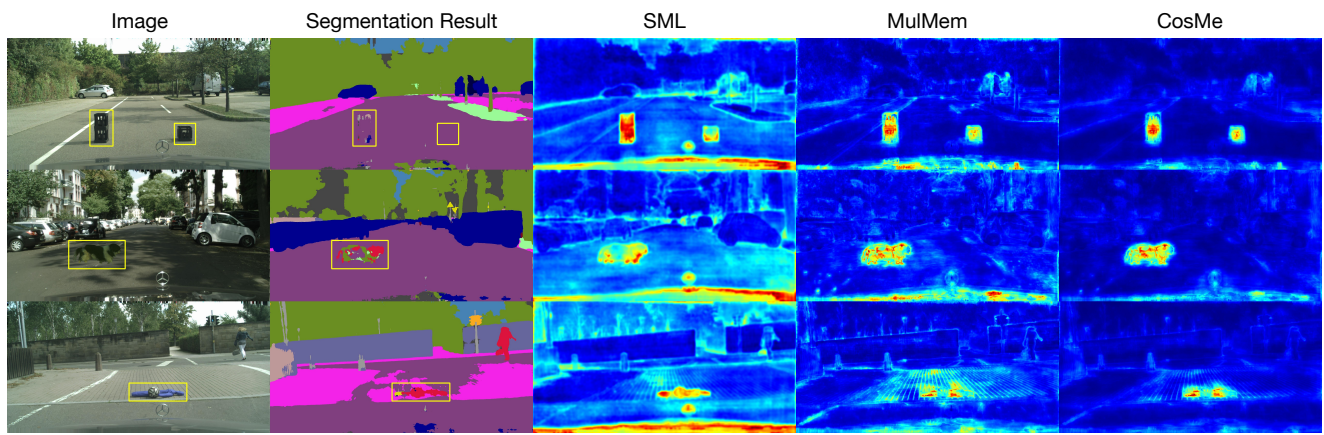}
    \captionof{figure}{\textbf{An overview of anomaly segmentation.} When encountered with anomalies, segmentation models may make mistakes, \emph{e.g.}, anomalous samples (marked by yellow boxes) are recognized as road, people and their mixture, which may cause accidents in autonomous driving. We show the anomaly segmentation results of standardized max logit (SML)~\cite{SML}, our memory baseline MulMem,and CosMe.}
    \label{fig:anomaly_scenarios}
    \vspace{-1mm}
\end{center}%
}]

\let\thefootnote\relax\footnote{$^{\textrm{\Letter}}$ Corresponding Author: \texttt{wei.shen@sjtu.edu.cn}}







\begin{abstract}
    \vspace{-14pt}
    Anomaly segmentation is a crucial task for safety-critical applications, such as autonomous driving in urban scenes, where the goal is to detect out-of-distribution (OOD) objects with categories which are unseen during training. The core challenge of this task is how to distinguish hard in-distribution samples from OOD samples, which has not been explicitly discussed yet. In this paper, we propose a novel and simple approach named \textbf{Co}nsensus \textbf{S}ynergizes with \textbf{Me}mory (CosMe) to address this challenge, inspired by the psychology finding that groups perform better than individuals on memory tasks. The main idea is 1) building a memory bank which consists of seen prototypes extracted from multiple layers of the pre-trained segmentation model and 2) training an auxiliary model that mimics the behavior of the pre-trained model, and then measuring the consensus of their mid-level features as complementary cues that synergize with the memory bank. CosMe is good at distinguishing between hard in-distribution examples and OOD samples. Experimental results on several urban scene anomaly segmentation datasets show that CosMe outperforms previous approaches by large margins.
\end{abstract}

\section{Introduction}
\label{sec:intro}

Semantic segmentation in urban scenes is an important technology for many vision-based applications. Current studies~\cite{fcn,unet,seg1,seg2,seg3,seg4} on semantic segmentation mainly focused on designing complex segmentation networks with higher segmentation capacities on in-distribution samples, while they paid less attention to out-of-distribution (OOD) samples, \emph{a.k.a}, anomalous objects, whose categories are unseen during training. A commonly-noticed shortcoming of current segmentation networks is that they are incapable of identifying anomalous objects. Instead, they can only predict an anomalous object as one seen category. This issue greatly impedes their uses in safety-critical applications such as autonomous driving in urban scenes. For example, the anomalous objects (marked by yellow boxes) in Fig.~\ref{fig:anomaly_scenarios} are predicted as a road by a segmentation network, which may lead to accidents. To address this issue, anomaly segmentation, a task to detect and segment out unseen anomalous objects, is attracting more and more attention.

\begin{figure}[t]
  \centering
    \includegraphics[width=0.95\linewidth]{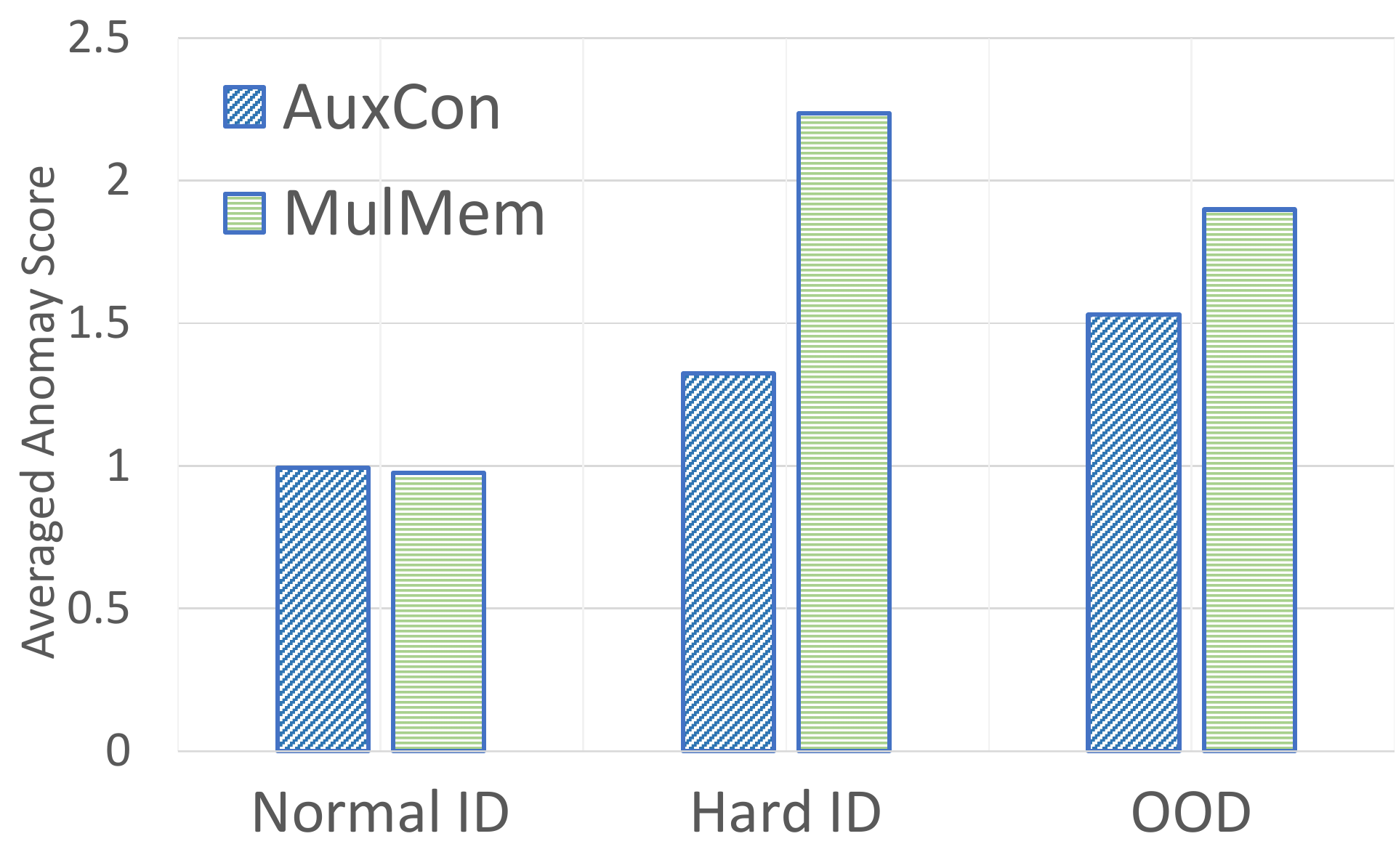}
   \caption{Statistical analysis of MulMem anomaly scores and AuxCon anomaly scores on Fishyscapes Lost \& Found. ``ID'' and ``OOD'' denote in-distribution samples and out-of-distribution samples, respectively. Normal in-distribution samples and hard in-distribution samples are distinguished by thresholding with the MulMem anomaly score at 95\% TPR. We can observe that the ability of MulMem to differentiate hard in-distribution samples from OOD samples is hardly guaranteed since the averaged MulMem anomaly score over hard in-distribution samples is even higher than that over OOD samples, while AuxCon shows a favorable ability to address this issue. }
   \vspace{-5mm}
   \label{fig:avg_ano}
\end{figure}

Previous approaches attempted to address this task by revealing anomalies according to prediction incorrectness of the segmentation networks, \emph{e.g.}, uncertainties over categories~\cite{MSP,SML} and re-synthesis errors caused by segmentation failures~\cite{imgr,SynthCP,noti}. However, they lack a mechanism to distinguish hard in-distribution samples from anomalies, and thus suffer from high false positive rates.

Differentiating between hard in-distribution samples and anomalies is the core challenge of anomaly segmentation. In this paper, we devote to addressing this challenge inspired by how humans deal with normal and hard examples: For a normal example, a human can confidently confirm his/her opinion about it is correct according to his/her memory, \emph{e.g.}, experience or knowledge; For a hard example, the human might be uncertain about it, but he/she can check with others and know his/her opinion is probably correct if others have consistent opinions. This is known as a general psychology finding that groups perform better than individuals on memory tasks~\cite{cac}.

Based on the above intuition, we introduce a novel and simple approach named \textbf{Co}nsensus \textbf{S}ynergizes with \textbf{Me}mory (CosMe) for anomaly segmentation. First, we present a strong memory-based baseline, named Multi-layer Memory (MulMem), which leverages a feature bank consisting of seen prototypes extracted from multiple layers of the pre-trained segmentation model to memorize seen objects with different scales. Then, we present a consensus-based module, named auxiliary consensus (AuxCon), in which an auxiliary network is trained to keep reaching a consensus with the pre-trained segmentation model in a self-supervised manner. This is achieved by explicitly maintaining hierarchical consistency between the auxiliary network and the pre-trained segmentation model on the feature representations of samples from seen categories.

Intuitively, whether an in-distribution sample is normal or hard can be determined by its distance to the prototypes in the MulMem feature bank. By this means, we divide the in-distribution samples into normal and hard sets. Then we compute the averaged anomaly scores for normal in-distribution samples, hard in-distribution samples and OOD samples according to MulMem (the minimum distance to prototypes) and AuxCon (the feature inconsistency), respectively. As illustrated in~\cref{fig:avg_ano}, the memory-based module can easily differentiate normal in-distribution samples from OOD samples, but its ability to distinguish hard in-distribution samples is hardly guaranteed. In contrast, the consensus-based module shows a clearly better differentiation ability between hard in-distribution samples and OOD samples than the memory-based module, while its overall discrimination between in-distribution samples and OOD samples is relatively smaller. These observations suggest good complementarity between MulMem and AuxCon. And thus, a simple combination of them, \emph{i.e.}, CosMe, has a strong ability for anomaly segmentation, especially can favorably distinguish hard in-distribution samples from anomalies.

Experimental results show that CosMe achieves consistent and substantial improvements over the state-of-the-art anomaly segmentation approaches on several urban scene datasets, and its performance is even on par with those methods using extra OOD samples for re-training.

In summary, our contributions are as follows:
\begin{itemize}
 \item
 We are the first to explicitly design a mechanism to tackle the challenge of hard in-distribution samples in anomaly segmentation.
 \item We propose CosMe, a novel approach which enjoys the benefits from the complementarity between memory-based prototype-level distance and feature-level inconsistency, with a good ability in differentiating between hard in-distribution samples and OOD samples.
 \item CosMe significantly outperforms the current state-of-the-art anomaly segmentation approach and even is comparable with the methods using extra OOD samples for re-training.
\end{itemize}


\section{Related Work}
The problem of detecting and segmenting unseen anomalous objects gradually attracts more and more attention. It is also related to out-of-distribution (OOD) detection. In this section, we first give a brief review of OOD detection, then describe previous approaches for anomaly segmentation in urban scenes.
\subsection{Out of Distribution (OOD) detection}

OOD detection is a broad concept, which is critical to ensuring the reliability of machine learning systems. There are many sub-tasks under this task, such as open set recognition~\cite{openset}, novelty detection~\cite{nd1,nd2}, \emph{etc}.  Since Hendryck and Gimpel~\cite{MSP} proposed the first OOD detection baseline in 2017, a plethora of approaches were developed, which can be categorized into clustering-based~\cite{clus,clus2}, uncertainty-based~\cite{streethazards,SML,odin,deepmetric}, reconstruction-based~\cite{memAE,imgr,SynthCP,synboost} and density-based~\cite{density1,density2}, \emph{etc}. As clustering-based is one type of most straightforward OOD detection method, we revisit it for anomaly segmentation.

\subsection{Anomaly Segmentation}

\subsubsection{Uncertainty-based}

Detecting anomalies based on model uncertainties is intuitive, since they are unseen during model training. Hendryck and Gimpel~\cite{MSP} proposed a baseline for OOD detection named ``maximum softmax probability'' (MSP), which measures anomaly scores by the maximum softmax probability outputted by the softmax classifier. Then they proposed ``maxlogit''~\cite{streethazards}. In maxlogit, the logits, \emph{i.e.}, the inputs of the softmax classifier are used as the anomaly scores instead. Jung \emph{et al.}~\cite{SML} proposed ``standardized maximum logit'' (SML) by improving ``maxlogit''. They used the statistics of the training set to standardize the ``maxlogit'' scores for each seen category, leading to a large improvement in anomaly segmentation results. However, these approaches lack a mechanism to distinguish hard in-distribution samples from anomalies. To address this challenge, some other approaches tried to first make the segmentation networks more sensitive to anomalous samples by either re-training them with a new loss function~\cite{deepmetric} or re-designing them by a new network architecture~\cite{Dense}, then applied the uncertainty measures. However, the segmentation networks modified by these strategies sacrifice their performance on seen categories. Chan \emph{et al.}~\cite{EntMax} utilized samples from the COCO dataset~\cite{coco} as OOD proxy for urban scenes and introduced an extra training objective to maximize the uncertainty on these samples. However, in practice, the categories of anomalous samples are unknown and inaccessible.


\subsubsection{Synthesis-based}
Recently, thanks to the rapid development of generative adversarial networks (GANs)~\cite{GAN,GauGAN}, one can reconstruct complex urban scenes from segmentation results. Some recent studies~\cite{imgr,SynthCP,noti} re-synthesized an image from a segmentation result and then compare it to the original input image to localize the anomalous instances. Synthesis-based approaches are unable to differentiate between hard in-distribution samples and OOD samples since segmentation results of hard samples are usually incorrect. In addition, their anomaly segmentation performance is heavily dependent on the generation quality of GANs and may suffer from degradation due to artifacts and style shifts in the synthesized images. Last, the serialized processing, \emph{i.e.}, segment-synthesize-compare, make this kind of approach difficult to be applied in practical real-time applications.

\subsubsection{Hybrid}
To achieve better anomaly segmentation results, Giancarlo \emph{et al.}~\cite{synboost} proposed a hybrid approach, which combines both uncertainty-based approach and synthesis-based approach. Nevertheless, their hybrid approach still suffers from the own issues of each component mentioned above. Besides, they implicitly made use of OOD samples to train a discriminator in their approach, which limits the generalization ability of their approach.





\section{Consensus Synergizes with Memory}

In this section, we first formulate the problem of anomaly segmentation, and then sketch out the overall framework of our method CosMe. Next, we describe the details of the two key components of CosMe, including the memory-based baseline \textbf{Multi-layer Memory (MulMem)} (\cref{sec:pm}) and the consensus-based module \textbf{Auxiliary Consensus (AuxCon)} (\cref{sec:cm}).

\begin{figure*}
  \centering
    \includegraphics[width=0.98\linewidth]{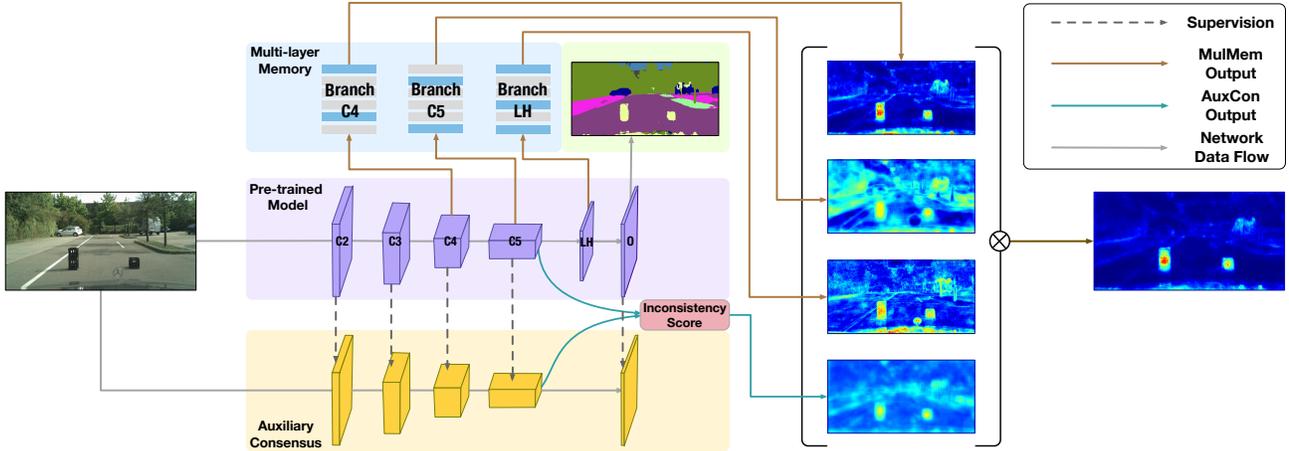}
    \caption{The overall framework of CosMe. It consists of two main modules: \textbf{Multi-layer Memory} (MulMem) and \textbf{Auxiliary Consensus} (AuxCon). Given a fixed pre-trained segmentation model, MulMem stores prototypes extracted from multiple layers ($\texttt{C2}-\texttt{C5}$ are Conv layers of the backbone, $\texttt{LH}$ and $\texttt{O}$ are the last hidden layer and  the last $1\times1$ Conv layer to compute the outputted logits over categories of the segmentation model, respectively) of the pre-trained segmentation model and AuxCon is an auxiliary model which is trained to maintain consistency with the segmentation model on in-distribution data. For the former, the distance to prototypes in MulMem is used as the MulMem anomaly score; For the latter, the inconsistency between the pre-trained model and the auxiliary model is used as the AuxCon anomaly score. These two kinds of anomaly scores     are combined to give the final anomaly prediction.
    }
  \label{fig:framework}
\end{figure*}

\subsection{Problem Setup}
\label{sec:setting}
We first set up the problem of anomaly segmentation.
Given a training set $\mathcal{T}=\{(\mathbf{x}^{(n)},\mathbf{y}^{(n)})_{n=1^N}$ for semantic segmentation with a category set $\mathcal{C}$, where $(\mathbf{x}^{(n)},\mathbf{y}^{(n)})$ denotes a pair of training image and its corresponding segmentation ground-truth and $y_{i,j}^{(n)}\in\mathcal{C}$ denotes the category label for the pixel at location $(i,j)$, a segmentation model $\mathbb{M}$ is pre-trained on $\mathcal{T}$, parameterized by $\bm{\Theta}$. Given a testing image $\mathbf{x}$ with categories from an unseen set $\mathcal{U}\cap\mathcal{C}=\emptyset$, the goal of anomaly segmentation is to segment out pixels that belong to unseen categories, based on $\mathbb{M}$ and $\mathcal{T}$. This can be achieved by assigning an anomaly score $\Upsilon_{i,j}(\mathbf{x})$ to each pixel $(i,j)$, so that
$$\min \Upsilon_{i,j}(\mathbf{x}), \text{if}~~y_{i,j} \in \mathcal{C};$$
\vspace{-6mm}
$$\max \Upsilon_{i,j}(\mathbf{x}), \text{if}~~y_{i,j} \in \mathcal{U}.$$


Some state-of-the-art anomaly segmentation approaches either retrain the pre-trained segmentation model or even make use of OOD data. However, we argue that these approaches have limitations, since 1) retraining may lead to negative effects on in-distribution segmentation performance; 2) the categories of OOD data are unknown and inaccessible in real-world applications.



\subsection{Overall Framework}
\label{sec:framework}
The overall framework of CosMe is shown in \cref{fig:framework}.
Except for the pre-trained segmentation model, there are mainly two parts in CosMe: one is the Multi-layer Memory (MulMem), the other is the Auxiliary Consensus (AuxCon). These two modules are combined to tackle the problem of hard in-distribution samples while fully utilizing the memories embedded in the segmentation model. We give a brief introduction to them as follows:


\begin{itemize}

    \item The \textbf{Mul}ti-layer \textbf{Mem}ory (MulMem) is a feature bank consisting of several feature sub-branches, each of which stores representative features, \emph{i.e.}, prototypes, outputted from a specific layer of the segmentation network. Whether a sample is anomalous is determined by the feature distance of the sample to the prototypes in sub-branches.

    \item The \textbf{Aux}iliary \textbf{Con}sensus (AuxCon) is an auxiliary model sharing information from the pre-trained model. Its task is to mimic the behavior of the pre-trained model on in-distribution data. When encountered with anomalies, the auxiliary model will show relatively big inconsistency with the pre-trained model. So the mimicking error of the auxiliary model is a kind of anomaly score naturally.

\end{itemize}

CosMe was built on the observation that hard in-distribution samples often lead to segmentation failures and are easily confused with anomalies. We are inspired by how humans distinguish uncertain samples. Humans tend to query others for suggestions when they cannot draw clear conclusions with their own memory. Thus, MulMem in CosMe imitates human memory, while AuxCon imitates someone else with similar experiences. For hard in-distribution samples, AuxCon can show relatively higher consistency than totally unseen anomalies.



\subsection{Multi-layer Memory}
\label{sec:pm}
Let $\mathbf{f}^{(l)}(\mathbf{x};\bm{\Theta})$ be the feature map for an input image $\mathbf{x}$, outputted from layer $l$ of the pre-trained segmentation model $\mathbb{M}$, and let $\mathbf{f}^{(l)}_{i,j}(\mathbf{x};\bm{\Theta})$ denote the feature vector at the location $(i,j)$ of this feature map. Our goal is to build a memory bank $\mathcal{M}=\{\mathcal{S}^{(l)}|l\in\mathcal{L}\}$ to store prototype features of seen samples from the training set $\mathcal{T}$, where $\mathcal{S}^{(l)}$ is a sub-branch of the bank to store prototype features outputted from layer $l$, and $\mathcal{L}$ is the set of layer of interests.

To memorize seen samples, a straightforward way is performing clustering on the training set to generate prototypes. Since the segmentation network processes data samples batch-wise, we propose a batch-based clustering algorithm to generate the prototypes. Without loss of generality, we describe our batch-based clustering algorithm by taking prototype generation for one feature sub-branch $\mathcal{S}^{(l)}$ as an example. We first set the feature sub-branch as an empty set: $\mathcal{S}^{(l)}=\{ \emptyset \}$, then we initialize this set by iteratively adding elements into it to form the prototypes. The elements are features from some randomly selected training images. Given a new element, \emph{e.g.}, $\mathbf{f}^{(l)}_{i,j}(\mathbf{x} ;\bm{\Theta})$, we first compute the cosine similarity $\phi(\cdot,\cdot)$ between the element $\mathbf{f}^{(l)}_{i,j}(\mathbf{x};\bm{\Theta })$ and each prototype $\mathbf{p} \in \mathcal{S}^{(l)}$ in current feature sub-branch, if this element is not similar to all the prototypes (determined by a similarity threshold $\tau$), then we add it to the feature sub-branch as a new prototype, \emph{i.e.}, $\mathcal{S}^{(l)} \leftarrow \mathcal{S}^{(l)} \cup \{\mathbf{f}^{(l)}_{i,j}(\mathbf{x};\bm{\Theta})\}$; otherwise, this element is not considered. This element adding process for prototype initialization is ended until the size of the feature sub-branch reaches a pre-set number $K$. The algorithm of prototype initialization for sub-branch $\mathcal{S}^{(l)}$ is shown in \cref{alg:PI}.

\begin{algorithm}[t!]
\caption{Memory Initialization for Sub-branch $\mathcal{S}^{(l)}$}
\label{alg:PI}
    \begin{algorithmic}[1]
    \REQUIRE Training batch $\mathcal{B}$, threshold $\tau$,  sub-branch size $K$
    \ENSURE Initialized $K$ prototypes $\mathcal{S}^{(l)}\leftarrow\{\mathbf{p}_k\}_{k=1}^K$
    \STATE $\mathcal{S}^{(l)} = \{\emptyset\}$
    \WHILE {$|\mathcal{S}^{(l)}|<K$}
    \STATE Randomly select an image $\mathbf{x}$ from $\mathcal{B}$
    \STATE Randomly select one $\mathbf{f}^{(l)}_{i,j}(\mathbf{x};\bm{\Theta})$ from its features
    \IF{$\max\{\phi\big(\mathbf{p},\mathbf{f}^{(l)}_{i,j}(\mathbf{x};\bm{\Theta})\big)|\forall\mathbf{p}\in\mathcal{S}^{(l)}\}<\tau$}
    \STATE   $\mathcal{S}^{(l)} \leftarrow \mathcal{S}^{(l)} \cup \{\mathbf{f}^{(l)}_{i,j}(\mathbf{x};\bm{\Theta})\}$
    \ENDIF
    \ENDWHILE
    \end{algorithmic}
\end{algorithm}

After prototype initialization, we learn the prototypes in $\mathcal{S}^{(l)}$ by a momentum update, given each training batch $\mathcal{B}$. Specifically, for each prototype $\mathbf{p}$ in $\mathcal{S}^{(l)}$, we update $\mathbf{p}$ by the features that are closest to $\mathbf{p}$, \emph{i.e.}, the highest cosine similarity. This can be achieved by maintaining a set $\mathcal{S}^{(l)}_p$ to store such features for prototype $\mathbf{p}$:
\begin{equation}
    \label{eq:update1}
    \mathcal{S}^{(l)}_p \leftarrow \{ \mathbf{f}\in\mathcal{F}_{\mathcal{B}} |\mathbf{p}=\arg\max_{\mathbf{p}'\in\mathcal{S}^{(l)}}\mathcal{Q}_{\mathcal{S}^{(l)}, \mathcal{F}_{\mathcal{B}}}\},
\end{equation}
where $\mathcal{F}_{\mathcal{B}} \leftarrow \{\mathbf{f}^{(l)}_{i,j}(\mathbf{x};\bm{\Theta})| \mathbf{x} \in \mathcal{B} \}$ is the set containing all the features of the images in batch $\mathcal{B}$ and $\mathcal{Q}_{\mathcal{S}^{(l)}, \mathcal{F}_{\mathcal{B}}}\leftarrow \{\phi(\mathbf{p},\mathbf{f})|\forall\mathbf{p}\in\mathcal{S}^{(l)},\forall\mathbf{f}\in\mathcal{F}_{\mathcal{B}}\}$  is the set containing all cosine similarities between each prototype $\mathbf{p}\in\mathcal{S}^{(l)}$ and each feature $\mathcal{F}_{\mathcal{B}}$. Finally, $\mathbf{p}$ is computed by a momentum update:
\begin{equation}
    \label{eq:update2}
    \mathbf{p} \leftarrow m \cdot \mathbf{p} + (1-m) \cdot \frac{1}{|\mathcal{S}^{(l)}_{p}|}\sum_{s^{(l)}_{p} \in \mathcal{S}^{(l)}_{p}} s^{(l)}_{p},
\end{equation}
where $m$ is a pre-defined momentum coefficient. The algorithm of prototype learning by the momentum update for sub-branch $\mathcal{S}^{(l)}$ is given in \cref{alg:MU}.



\begin{algorithm}[ht!]
\caption{Memory Learning for Sub-branch $\mathcal{S}^{(l)}$}
\label{alg:MU}
    \begin{algorithmic}[1]
    \REQUIRE Training batch $\mathcal{B}$, coefficient $m$, initialized $\mathcal{S}^{(l)}$
    \ENSURE Updated $K$ prototypes $\mathcal{S}^{(l)} =\{\mathbf{p}_k\}_{k=1}^K$
        \STATE $\mathcal{F}_{\mathcal{B}} \leftarrow \{\mathbf{f}^{(l)}_{i,j}(\mathbf{x};\bm{\Theta})| \mathbf{x} \in \mathcal{B} \}$
        \STATE $\mathcal{Q}_{\mathcal{S}^{(l)}, \mathcal{F}_{\mathcal{B}}}\leftarrow \{\phi(\mathbf{p},\mathbf{f})|\forall\mathbf{p}\in\mathcal{S}^{(l)},\forall\mathbf{f}\in\mathcal{F}_{\mathcal{B}}\}$
        \FOR {$\mathbf{p} \in \mathcal{S}^{(l)}$}
            \STATE $\mathcal{S}^{(l)}_p \leftarrow \{ \mathbf{f}\in\mathcal{F}_{\mathcal{B}} |\mathbf{p}=\arg\max_{\mathbf{p}'\in\mathcal{S}^{(l)}}\mathcal{Q}_{\mathcal{S}^{(l)}, \mathcal{F}_{\mathcal{B}}}\}$
            \STATE $\mathbf{p} \leftarrow m \cdot \mathbf{p} + (1-m) \cdot\frac{1}{|\mathcal{S}^{(l)}_{p}|} \sum_{s^{(l)}_{p} \in \mathcal{S}^{(l)}_{p}} s^{(l)}_{p}$
        \ENDFOR
    \end{algorithmic}
\end{algorithm}

With the learned feature sub-branch $\mathcal{S}^{(l)}$, given an input image $\mathbf{x}$, an anomaly score map $\bm{\gamma}^{(l)}(\mathbf{x})$ for this input image is computed by
\begin{equation}
    \label{eq:as}
    \gamma_{i,j}^{(l)} = 1 - \max\{\phi\big(\mathbf{p},\mathbf{f}^{(l)}_{i,j}(\mathbf{x};\bm{\Theta})\big)|\forall\mathbf{p}\in\mathcal{S}^{(l)} \},
\end{equation}
where $\gamma^{(l)}_{i,j}(\mathbf{x})$ denotes the anomaly score at the location $(i,j)$ of the anomaly score map $\bm{\gamma}^{(l)}(\mathbf{x})$.

Since several sub-branches form the feature bank $\mathcal{M}=\{\mathcal{S}^{(l)}|l\in\mathcal{L}\}$ of MulMem, we compute the anomaly score map ${\Gamma}(\mathbf{x})$ given by the feature bank $\mathcal{M}$ by a simple combination of the anomaly score maps given by each sub-branch:
\begin{equation}
    \Gamma_{i,j}(\mathbf{x}) = \Pi_{l\in \mathcal{L}}{\gamma}_{i,j}^{(l)}(\mathbf{x}),
\end{equation}
where $\Gamma_{i,j}(\mathbf{x})$ is the MulMem anomaly score at the location $(i,j)$ of the anomaly score map ${\bm{\Gamma}}(\mathbf{x})$. We additionally adopt the standardization strategy in~\cite{SML} to normalize the MulMem anomaly scores in ${\bm{\Gamma}}(\mathbf{x})$.

\subsection{Auxiliary Consensus}
\label{sec:cm}

As shown in \cref{fig:framework}, the auxiliary consensus module explicitly ensures feature consistency between the pre-trained segmentation model and an auxiliary model. This is achieved by self-supervised learning without using any segmentation annotations of the training set $\mathcal{T}$. Given the pre-trained segmentation model $\mathbb{M}$, we build an auxiliary model $\mathbb{M}^{\prime}$ parameterized by $\bm{\Theta}^{\prime}$ which has the same down-sampling schedule as $\mathbb{M}$, so that for a layer $l$ in $\mathbb{M}$ we can find its corresponding layer $l^{\prime}$ in $\mathbb{M}^{\prime}$. For example, ResNet50~\cite{resnet} can be the backbone of an auxiliary model for a segmentation model with ResNet101 as the backbone.

Let $\mathcal{L}_s$ be a set of layers of $\mathbb{M}$, (\emph{e.g.}, for ResNet, $\mathcal{L}_s$ can be the last Conv layers of the five Conv blocks $\mathcal{L}_s=\{\texttt{C1},\texttt{C2},\texttt{C3},\texttt{C4},\texttt{C5}\}$) which is used to supervise the corresponding layers of $\mathbb{M}^{\prime}$. For each layer $l\in\mathcal{L}_s$, let $s^{(l)}$ be the size of the feature map $\mathbf{f}^{(l)}(\mathbf{x};\bm{\Theta})$, and $l^{\prime}$ is its corresponding layer in $\mathbb{M}^{\prime}$, our learning purpose is to enforce the feature map $\mathbf{g}^{(l^{\prime})}(\mathbf{x};\bm{\Theta}^{\prime})$ outputted by layer $l^{\prime}$ of $\mathbb{M}^{\prime}$ to approach $\mathbf{f}^{(l)}(\mathbf{x};\bm{\Theta})$.

Towards this end, we fix $\bm{\Theta}$, and minimize the following loss function on the training set $\mathcal{T}$:

\begin{equation}
    \label{eq:sup}
    L=\sum_{\mathbf{x}\in\mathcal{T}}\sum_{l\in \mathcal{L}_s}  \frac{1}{s^{(l)}}|| \mathbf{f}^{(l)}(\mathbf{x}, \bm{\Theta}) - \mathbf{g}^{(l^{\prime})}(\mathbf{x};\bm{\Theta}^{\prime}) ||_F^2,
\end{equation}
where $||\cdot||_F$ means the Frobenius norm of a matrix. The overall training algorithm is shown by \cref{alg:AML}.


During inference, to compute anomaly scores, we select a subset $\mathcal{L}_e$ from $\mathcal{L}_s$ as an evaluation set. Given a testing image $\mathbf{x}$, the anomaly score $\psi_{i,j}^{(l)}$ at each location $(i,j)$ is computed by:
\begin{equation}
    \psi_{i,j}^{(l)} = \frac{1}{C^{(l)}}||\mathbf{f}^{(l)}_{i,j}(\mathbf{x}, \bm{\Theta}) - \mathbf{g}^{(l^\prime)}_{i,j}(\mathbf{x}, \bm{\Theta^\prime})||_2^2,
\end{equation}
where $||\cdot||_2$ means $\ell_2$ norm and $C^{(l)}$ denotes the dimension of the feature channel. Then AuxCon anomaly score at each location $(i,j)$ is:
\begin{equation}
    \Psi_{i,j}(\mathbf{x}) = \Pi_{l \in \mathcal{L}_e} \psi_{i,j}^{(l)}(\mathbf{x}).
\end{equation}
Finally, the CosMe anomaly score $\Upsilon_{i,j}(\mathbf{x})$ is calculated by:
\begin{equation}
    \Upsilon_{i,j}(\mathbf{x}) = \Psi_{i,j}(\mathbf{x}) \cdot \Gamma_{i,j}(\mathbf{x}).
\end{equation}




\begin{algorithm}[t!]
\caption{Auxiliary Model Learning}
\label{alg:AML}
    \begin{algorithmic}[1]
    \REQUIRE Training set $\mathcal{T}$, layer set $\mathcal{L}_s$, learning rate $\eta$
    \ENSURE The auxiliary model parameterized $\bm{\Theta}'$

        \STATE Initialize $\bm{\Theta}^\prime$ randomly
        \FOR {each image batch $\mathcal{B}\subset\mathcal{T}$}

            \STATE \mbox{$L \leftarrow \sum_{\mathbf{x}\in\mathcal{B}}\sum_{l\in \mathcal{L}_s}  \frac{1}{s^{(l)}}|| \mathbf{f}^{(l)}(\mathbf{x}, \bm{\Theta}) - \mathbf{g}^{(l^\prime)}(\mathbf{x}, \bm{\Theta^\prime}) ||_F^2$}

            \STATE $\bm{\Theta}^\prime\leftarrow \bm{\Theta}^\prime - \eta \cdot \frac{\partial L}{\partial \bm{\Theta}^\prime}$
        \ENDFOR
    \end{algorithmic}
\end{algorithm}

\section{Experiments}

In this section, we describe the datasets used for our experiments, implementation details, evaluation metrics and experimental results.
\subsection{Datasets}
We conduct our experiments on four widely-used anomaly segmentation datasets: Fishyscapes Lost \& Found~\cite{fishyscapes}, Fishyscapes Static~\cite{fishyscapes}, Road Anomaly~\cite{imgr} and Streethazards~\cite{streethazards}.

\label{sec:lf}
\noindent\textbf{Fishyscapes Lost \& Found.} Fishyscapes Lost (FS) \& Found~\cite{fishyscapes} is a high-quality image dataset containing real obstacles on roads. Based on the original Lost \& Found~\cite{l_f} dataset, the FS Lost \& Found dataset also follows the same setup as Cityscapes\cite{cityscapes}, which is a widely used dataset in urban-scene segmentation. It contains real urban images with 37 types of unexpected road obstacles and 13 different street scenarios (\emph{e.g.}, different road surface appearances, strong illumination changes, etc). FS Lost \& Found includes a public validation set of 100 images and a hidden test set of 275 images for the benchmarking.

\label{sec:static}
\noindent\textbf{Fishyscapes Static.} Fishyscapes (FS) Static~\cite{fishyscapes} is built based on validation set of Cityscapes\cite{cityscapes}. Anomalous objects collected from PASCAL VOC~\cite{voc} are superimposed on Cityscapes seamlessly to ensure matching with the style of Cityscapes. This dataset contains a publicly available validation set with 30 images and a test set hidden for benchmarking with 1,000 images.

\label{sec:ra}
\noindent\textbf{Road Anomaly.} Road Anomaly~\cite{imgr} captures dangerous scenes where vehicles encounter on roads. It consists of 60 images collected from the Internet, including strange objects on roads (\emph{e.g.}, animals, rocks, etc.), with a resolution of 1280 × 720. Since this dataset is not collected under conditions similar to Cityscapes, there is a big domain gap between them. This dataset can be used to verify the generalization ability of an anomaly segmentation approach.

\label{sec:streethazards}
\noindent\textbf{Streethazards.} Streethazards~\cite{streethazards} is an anomaly segmentation dataset created by using the Unreal Engine along with the CARLA simulation environment. This dataset contains 5125 image and semantic segmentation ground-truth pairs for training, 1,031 pairs without anomalies for validation, and 1,500 test pairs with anomalies. There are 250 unique anomaly models of diverse types in total and 12 classes of objects used for training.
\subsection{Implementation Detail}
\label{sec:impd}
\noindent\textbf{Pre-trained model.} For fair comparison, we follow~\cite{streethazards,deepmetric,SynthCP} to adopt PSPnet\cite{PSP} with ResNet101~\cite{resnet} as the segmentation model on Streethazards and follow~\cite{synboost}\cite{SML} to adopt DeepLabV3+\cite{deeplabv3plus} with ResNet101 as the segmentation model on the other three datasets.

\noindent\textbf{Multi-layer Memory.} We maintain three memory sub-branches, in which prototypes are extracted from the outputs of $\texttt{C4}$ layer and $\texttt{C5}$ layer of the ResNet101 backbone as well as the last hidden layer ($\texttt{LH}$) of the segmentation model. The similarity threshold $\tau = 0.85$.


\noindent\textbf{Auxiliary Consensus.} We use ResNet50 as the backbone of the auxiliary model. The supervision layer set is $\mathcal{L}_s = \{\texttt{C2},\texttt{C3},\texttt{C4},\texttt{C5},\texttt{LH},\texttt{O}\}$, where $\texttt{O}$ is the output layer of the segmentation model, \emph{i.e.}, the last $1\times 1$ Conv layer to compute the outputted logits over categories. The evaluation layer set for computing AucCon anomaly score is $\mathcal{L}_e = \{\texttt{C5}\}$.

\subsection{Evaluation Metrics}
Following~\cite{streethazards, SML, SynthCP}, three metrics are used for evaluation: area under receiver operating curve (\textbf{AUROC}), false positive rate at 95\% true positive rate (\textbf{FPR95}) and average precision (\textbf{AP}). Since anomalous samples are much less than in-distribution samples, the data imbalance suggests FPR95 and AP are major evaluation metrics.

\subsection{Comparison with Previous Approaches}

\begin{table*}[ht!]
  \centering
  \begin{tabular}{l|c|c|cc|cc}
    \toprule
    \multirow{2}*{Method} &  \multirow{2}{*}{\shortstack{Utilizing\\OOD Data}}& \multirow{2}{*}{\shortstack{Requiring\\re-training}} & \multicolumn{2}{c}{FS Lost \& Found} & \multicolumn{2}{|c}{FS Static} \\
    \cline{4-7} & & & FPR95 $\downarrow$ & AP $\uparrow$ & FPR95 $\downarrow$ & AP $\uparrow$\\
    \midrule
    \midrule
    {MSP}~\cite{MSP}& \xmark &\xmark & 44.85 & 1.77 & 39.83 & 12.88 \\
    {Entropy}~\cite{MSP}& \xmark &\xmark & 44.83 & 2.93 & 39.75 & 15.41 \\
    {Density - Single-layer NLL}~\cite{fishyscapes}& \xmark &\xmark & 32.90 & 3.01 & 21.29 & 40.86  \\
    {kNN Embedding - density}~\cite{fishyscapes} & \xmark &\xmark & 30.02 & 3.55 & 20.25 & 44.03  \\
    {Density - Minimum NLL}~\cite{fishyscapes}& \xmark &\xmark & 47.15 & 4.25 & 17.43 & 62.14  \\
    {Image Resynthesis}~\cite{imgr}& \xmark &\xmark & 48.05 & 5.70 & 27.13 & 29.60  \\
    {SML}~\cite{SML} & \xmark &\xmark & 21.52 & 31.05 & 19.64 & 53.11  \\
    \midrule
    {Synboost}~\cite{synboost} & \cmark &\xmark & 15.79 & \textbf{43.22} & 18.75 & 72.59  \\
    {Density - Logistic Regression}~\cite{fishyscapes} & \cmark &\cmark & 24.36 & 4.65 & 13.39 & 57.16  \\
    {Bayesian Deeplab}~\cite{bd} & \xmark &\cmark & 38.46 & 9.81 & 15.50 & 48.70  \\
    {OoD Training - Void Class}~\cite{fishyscapes}& \cmark &\cmark & 22.11 & 10.29 & 19.40 & 45.00  \\
    {Discriminative Outlier Detection Head}~\cite{Dense} & \cmark &\cmark & 19.02 & 31.31 & \textbf{0.29} & \textbf{96.76}  \\
    {Dirichlet Deeplab}~\cite{diric}& \cmark &\cmark & 47.43 & 34.28 & 84.60 & 31.3  \\
    \midrule
    {CosMe (Ours)} & \xmark &\xmark & \textbf{13.32} & \textbf{41.95} & \textbf{5.74} & \textbf{69.72}\\

    \bottomrule
  \end{tabular}
  \caption{Comparison with previous approaches reported in Fishyscapes Leaderboard \protect\footnotemark. The top part of the table shows the approaches with the same setting as our CosMe, \emph{i.e.}, no re-training and no extra OOD data. The bottom part shows the approaches which require retraining or extra OOD data. Our method outperforms the approaches with the same setting and even most of the approaches with re-training or extra OOD data, by large margins.}
  \label{tab:FS_leaderboard}
  \vspace{-2mm}
\end{table*}
\footnotetext{\url{https://fishyscapes.com/results}}

\noindent\textbf{Fishyscapes test sets.}
We first compare CosMe with other anomaly segmentation approaches on Fishyscapes (FS Lost \& Found and FS Static) test sets. Note that, Fishyscapes test sets are private. We get the results from Fishyscapes Leaderboard. According to \cref{tab:FS_leaderboard}, we achieve a new SOTA performance compared with approaches without re-training or extra OOD data and outperform them by large margins. Moreover, our performance is even better than most of the approaches with re-training or extra OOD data. CosMe is on par with Synboost, which is the SOTA with re-training on FS Lost \& Found. Compared with Synboost, which needs to re-synthesis images from the segmentation label and then compare them with the original input images, calculation in our approaches can be parallelized. Once training is finished, there is no dependency between the pre-trained model and the auxiliary model. This merit makes our model more suitable for practical applications.

 \begin{table*}[ht!]
  \centering
  \setlength{\tabcolsep}{1.85mm}{
  \begin{tabular}{l|ccc|ccc|ccc}
    \toprule
    \multirow{2}*{Method} & \multicolumn{3}{c}{FS Lost \& Found} & \multicolumn{3}{|c}{FS Static}& \multicolumn{3}{|c}{Road Anomaly}\\
    \cline{2-10} & FPR95 $\downarrow$ & AUROC $\uparrow$ & AP $\uparrow$ & FPR95 $\downarrow$ & AUROC $\uparrow$ & AP $\uparrow$& FPR95 $\downarrow$ & AUROC $\uparrow$ & AP $\uparrow$ \\
    \midrule
    \midrule
    {MSP}~\cite{MSP} & 45.63 & 86.99 & 6.02 & 34.10 & 88.94 & 14.24 & 68.44 & 73.76 & 20.59\\
    {MaxLogit}~\cite{streethazards} & 38.13 & 92.00 & 18.77 & 28.50 & 92.80 & 27.99 & 64.85 & 77.97 & 24.44\\
    {SynthCP}~\cite{SynthCP} & 45.95 & 88.34 & 6.54 & 34.02 & 89.90 & 23.22 & 64.69 & 76.08 & 24.87 \\
    {SML}~\cite{SML} & 14.53 & 96.88 & 36.55 & 16.75 & 96.69 & 48.67 & \textbf{49.74} & 81.96 & 25.82 \\
    \midrule
    {$\rm LDN\_BIN^{{\color{cyan}{\P}},{\color[RGB]{246,124,124}{\sharp}}}$}~\cite{Dense} & 23.97 & 95.59 & 45.71 & - & - & - & - & - & - \\

    \midrule
    {MulMem} (Ours) &14.47 & 97.39 & 41.73 & 5.07 & 98.87 & 65.61 & 63.38 & 80.02 & 29.49 \\
    AuxCon (Ours)   & 18.68 & 95.79 & 19.52 & 5.45 & 98.76 & 62.01 & 51.33 & 84.61 & 38.40  \\
    {CosMe (Ours)} & \textbf{11.65} & \textbf{98.11} & \textbf{50.22} & \textbf{1.47} & \textbf{99.58} & \textbf{79.25} &  51.04 & \textbf{85.92} &  \textbf{41.11} \\

    \bottomrule
  \end{tabular}
  }
  \caption{Comparison on Fishyscapes validation sets and Road Anomaly. $\color{cyan}{\P}$ and $\color[RGB]{246,124,124}{\sharp}$ indicate re-training and extra OOD data, respectively.}
  \label{tab:fishyscapes_val}
  \vspace{-2mm}
\end{table*}

\noindent\textbf{Fishyscapes validation sets.}
\label{sec:fvs}
\cref{tab:fishyscapes_val} shows our comparison results on the Fishyscapes (FS Lost \& Found and FS Static) validation sets. The results on Fishyscapes validation set demonstrate that both MulMem and CosMe outperform the previous approaches without re-training or extra OOD data by large margins. Especially, CosMe even outperforms LDN\_BIN, which utilized OOD data to retrain their segmentation model. Compared with SML~\cite{SML}, the previous SOTA approach without re-training or extra OOD data, \textbf{the major improvement of CosMe on FS Lost \& Found comes from AuxCon}, while \textbf{the improvement on FS Static is mainly thanks to MulMem}. This phenomenon further confirms the improvement in CosMe are mainly comes from tackling hard in-distribution samples, since there are more hard in-distribution samples in FS lost \& Found than FS Static: As we introduced in \cref{sec:lf}, the images in FS Lost \& Found come from real driving scenes, while the anomalous samples in FS Static are cut and pasted from other datasets, such as PASCAL VOC. The domain gap between normal background (Cityscapes) and anomalous foreground (PASCAL VOC) in FS Static can benefit memory-based anomaly segmentation, which relatively reduces the difficulty of segmentation.


\begin{table}[ht!]
  \centering
  \setlength{\tabcolsep}{1.6mm}{
  \begin{tabular}{l|ccc}
    \toprule
    Method & FPR95 $\downarrow$ & AUROC $\uparrow$ & AP $\uparrow$ \\
    \midrule
    \midrule
    {MSP}~\cite{MSP} & 33.7 & 87.7 & 6.6\\
    {SynthCP}~\cite{SynthCP} & 28.4 & 88.5 & 9.3\\
    {MaxLogit}~\cite{streethazards} & 26.5 & 89.3 & 10.6\\
    \midrule
    {$\rm Dropout^{\color{cyan}{\P}}$}~\cite{dropout} & 79.4 & 69.9 & 7.5 \\
    {$\rm DML^{\color{cyan}{\P}}$ }~\cite{deepmetric} & 17.3 & 93.7 & 14.7\\
    {$\rm LDN\_BIN^{{\color{cyan}{\P}},{\color[RGB]{246,124,124}{\sharp}}}$}~\cite{Dense} & 30.9 & 89.7 & 18.8\\
    \midrule
     CosMe (PSPnet) & 23.2 & 91.3 & 16.8\\
    {CosMe (DeepLabV3+)} & \textbf{15.5} & \textbf{94.6} & \textbf{19.7} \\
    \bottomrule
  \end{tabular}
  }
  \caption{Comparison results on Streethazards. $\color{cyan}{\P}$ and $\color[RGB]{246,124,124}{\sharp}$ indicate re-training and extra OOD data, respectively.}
  \label{tab:streethazards}
  \vspace{-2mm}
\end{table}

\noindent\textbf{Road Anomaly.}
\label{sec:RoadAnomaly}
As shown in \cref{tab:fishyscapes_val}, our results on Road Anomaly are significantly better than others and on par with SML~\cite{SML}. Since there is a big domain gap between Road Anomaly and Cityscapes, there are massive hard in-distribution samples in Road Anomaly. In this case,  AuxCon helps to improve performance much more than MulMem, which further evidences that AuxCon is adept at detecting hard in-distribution samples.

\noindent\textbf{Streethazards.}
\cref{tab:streethazards} shows the results on the test set of Streethazards. CosMe outperforms the previous approaches without re-training or extra OOD data by large margins and is on par with DML and LDN\_BIN which require re-training. Note that when we replace PSPnet with DeepLabV3+ as our pre-trained segmentation model, the performance is further improved and outperforms LDN\_BIN. This result implies the memories embedded in more powerful models might be stronger.

\subsection{Ablation Study}
We have shown ablation results w.r.t. the two sub-modules of CosMe in ~\cref{tab:fishyscapes_val}. Now, we conduct our ablation study w.r.t. the design in each sub-module on the FS Lost \& Found validation set.


\noindent \textbf{Ablation on layer set $\mathcal{L}$ for sub-branches.} $\mathcal{L}$ is introduced in~\cref{sec:pm}, which contains the layers used for prototype learning in the  memory bank. The ablation result is shown in the upper part of~\cref{tab:Li}. Note that this ablation is done solely on MulMem, without the help of AuxCon. It shows that when $\mathcal{L} = \{\texttt{C4},\texttt{C5},\texttt{LH}\}$, MulMem achieves the best performance. This evidences there are rich memories embedded in the segmentation network, which are not fully exploited by the previous approaches.

\begin{table}[t!]
  \centering
  \setlength{\tabcolsep}{1.6mm}{
  \begin{tabular}{c|c|ccc}
    \toprule
    & Layers in Set & FPR95 $\downarrow$ & AUROC $\uparrow$ & AP $\uparrow$ \\
    \midrule
    \midrule
    \multirow{5}*{$\mathcal{L}$} & $\{\texttt{C4}\}$ & 34.55 & 93.17 & 9.81\\
    &$\{\texttt{C5}\}$ & 36.49 & 92.97 & 22.11\\
    &$\{\texttt{LH}\}$ & 20.84 & 95.74 & 12.32\\
    &$\{\texttt{C4},\texttt{C5}\}$ & 23.94 & 95.34 & 27.38\\
    \rowcolor{gray!15}&$\{\texttt{C4},\texttt{C5},\texttt{LH}\}$ & \textbf{14.47} & \textbf{97.39} & \textbf{41.73}\\
    \midrule
    \multirow{7}*{$\mathcal{L}_e$} & $\{\texttt{C4}\}$ & 10.58 & 98.13 & 44.99\\
    \rowcolor{gray!15}&$\{\texttt{C5}\}$ & 11.65 & 98.11 & \textbf{50.22}\\
    &$\{\texttt{LH}\}$ & 15.45 & 97.04 & 35.91\\
    &$\{\texttt{O}\}$ & 16.90 & 96.50 & 27.95\\
    &$\{\texttt{C4},\texttt{C5}\}$ & \textbf{10.39} & \textbf{98.24} & 49.96\\
    &$\{\texttt{C4},\texttt{C5},\texttt{LH}\}$ & 13.09 & 97.67 & 42.74\\
    &$\{\texttt{C4},\texttt{C5},\texttt{LH},\texttt{O}\}$ & 15.07 & 97.00 & 32.28\\
    \bottomrule
  \end{tabular}
  }
  \caption{Ablation results for the selection of $\mathcal{L}$ (upper part) and $\mathcal{L}_e$ (lower part).}
  \label{tab:Li}
  \vspace{-4mm}
\end{table}

\noindent \textbf{Ablation on evaluation layer set $\mathcal{L}_e$.}  By fixing $\mathcal{L} = \{\texttt{C4},\texttt{C5},\texttt{LH}\}$, we then conduct ablation on evaluation layer set $\mathcal{L}_e$. As shown in the lower part of \cref{tab:Li}, when $\mathcal{L}_e = \{\texttt{C5}\}$, CosMe reaches its best performance.




\section{Discussion}
Compared with previous classic OOD detection approaches, maxlogit~\cite{streethazards} and MSP~\cite{MSP}, we measure anomalies in a white box manner, with a requirement to access the internal structure of the pre-trained segmentation model. But, here we give an intuitive explanation for the necessity of the access: According to Information Bottleneck (IB) theory~\cite{infobott}, the calculation process of neural networks can be seen as a kind of filtering process. Redundant information is filtered by the multi-layer architectures of deep networks. Since the optimization goal of training a segmentation model is to maximize the model's prediction toward ground-truth in-distribution categories, information for OOD detection is filtered to a certain extent. In summary, information only from the final prediction is not sufficient for anomaly segmentation.


\section{Conclusion}
In this paper, we pointed out the core challenge of anomaly segmentation is the existence of hard in-distribution samples. Based on the psychology finding of consensus processes in group recognition memory performance, we proposed ``Consensus Synergizes with Memory'' (CosMe), which utilizes inconsistency with an auxiliary model to complement the memory-based prototype-level distance for anomaly segmentation. 
Our approach was verified on various datasets and achieved superior results on all of them.
Note that, our approach has no constraint on the segmentation network and can be parallelized. This merit shows its potential in practical applications.

{\small
\bibliographystyle{ieee_fullname}
\bibliography{egbib}
}

\end{document}